\documentclass[letterpaper]{article}

\usepackage{amsmath}
\usepackage{amsfonts}

\usepackage{graphicx}
\usepackage{caption}
\usepackage{subcaption}

\usepackage{aaai}
\usepackage{times}
\usepackage{helvet}
\usepackage{courier}

\DeclareMathOperator*{\argmin}{arg\,min}
\DeclareMathOperator*{\argmax}{arg\,max}

\title{A genetic algorithm for autonomous navigation in partially observable domain}
\author{
	Maxim Borisyak, Andrey Ustyuzhanin\\
    Moscow Institute of Physics and Technology
}

\begin{document}
%

\maketitle

\begin{abstract}
\begin{quote}
	The problem of autonomous navigation is one of the basic problems for robotics.
    Although, in general, it may be challenging when an autonomous vehicle is placed
    into partially observable domain.
    In this paper we consider simplistic environment model and introduce a navigation algorithm
    based on Learning Classifier System.
\end{quote}
\end{abstract}

\section{Introduction}
The problem of navigation in partially observable domain may be computationally challenging since
state space of an environment in general grows exponentially with the size of the environment. Current studies suggest mostly usage of Partially Observable Markov Decision Process, for example, \cite{koenig1998xavier}, \cite{simmons1995probabilistic}, \cite{cassandra1994acting} however POMDP usually implies
computational challenges \cite{papadimitriou1987complexity} which make direct application quite difficult.
To avoid this problems number of technique is used such as division of domain state space \cite{dean1993planning}, hierarchical POMDP \cite{foka2002predictive} etc.

Alternative approaches also take place, including fuzzy logic \cite{pratihar1999genetic}, `bug' algorithms \cite{zohaib2013intelligent}. It has been recently shown that reactive navigation models can be successfully trained for the problem of obstacle avoidance \cite{ram1994using}, \cite{whitbrook2008genetic}.
In this paper we will continue genetic approach to train models for navigation in partially observable domains.

In contrast to mentioned works to avoid unnecessary complications we consider simplistic model of domain.
An autonomous robot is placed into cellular two-dimensional static environment with fixed width and height ($W$ and $H$), each cell of which is either occupied or free:
$$U \subseteq \{-1, 1\}^{W \times H}$$
where $U$ --- predefined set of possible domains; $-1$ corresponds to the free state of a cell, $1$ --- occupied.

The robot is allowed to occupy exactly one free cell
(which determines its position), has one of four directions and can execute 3 commands:
go forward (according to its direction) and change its direction by turning left or right, which forms set of possible actions $\mathcal{A}$. Robot's goal is to find sequence of commands to reach predefined cell with coordinates $(x^f, y^f)$.

The robot has only one sensor, vision that reveals state of cells in fixed radius circle around its position, and also adds unknown cell state (which is encoded as $0$):
$$V = \{-1, 0, 1\}^{W \times H}$$
where $V$ --- set of possible visual observations.

If vision function is $v^r$ where $r$ is vision radius then observation from position $(x_0, y_0)$ is:
\begin{align*}
	v^r(x_0, y_0) &= v \in V;\\
    v(x, y) &= \begin{cases}
    	u(x, y),& \mathrm{if}\, (x_0 - x)^2 + (y_0 - y)^2 \leq r;\\
        0, & \mathrm{otherwise}
	\end{cases}
\end{align*}
where $u \in U$ corresponds to the current domain.

To avoid unnecessary quality loss robot receives accumulated map of the domain where all observations are imposed, which can also be expressed by an element of $V$.

Set of robot's observations $\mathcal{O}$ consists of current vision $V$, position $\mathbb{N}^2$, direction $D = \{ (x, y) \mid x, y \in \{-1, 0, 1\}, |x| + |y| = 1 \}$ and goal $\mathbb{N}^2$:
\begin{align}
	\mathcal{O} = V \times \mathbb{N}^2 \times D \times \mathbb{N}^2
\end{align}

We define robot's navigation policy $\pi$ as a function $\psi$,
some state space $S_{\psi}$ and initial state $s_0 \in S_{\psi}$:
\begin{align}
	\psi(\omega^t) = (s^{t + 1}, a) \label{eq:policy}
\end{align}
where $t$ is number of current step, $\omega^t$ --- current observation,
$s^t, s^{t + 1} \in S_{\psi}$ --- policy states before and after making decision, $a \in \mathcal{A}$ --- the result action.

By executing policy $\pi$ on each step robot forms its path $p$ in the domain $u \in U$. Quality of a policy in particular domain $u$ is measured by the following cost function:
\begin{align}
	C_u(p) = \frac{|p|}{|p^*_u|} \label{eq:cost}
\end{align}
where $p^*_u$ is the shortest path from starting point to the goal in domain $u$.

It can be shown that Pareto optimal policy $\pi^*$ either
doesn't exist or the optimal policy reaches cost limit:
\begin{align*}
\left[ \forall \pi: \forall u \in U: C(p^{\pi^*}_u) \leq C(p^{\pi}_u)\right] \Longleftrightarrow\\
 \left[ \forall u \in U: C(p^{\pi^*}_u) = 1 \right]
\end{align*}
where $p^\pi_u$ --- path produced by policy $\pi$.

But since the last case where Pareto optimal policy exists is not a general case,
following POMDP approach, we introduce probability space $\Omega = (U, 2^U, P)$, where
$P(u)$ corresponds to the prior probability
of robot being placed in the domain $u \in U$, and corresponding cost function\footnotemark:
\begin{align}
	C(\pi) = \mathbb{E}_u \frac{|p^\pi_u|}{|p^*_u|} \label{eq:quality}
\end{align}
Note that here expectation operator $\mathbb{E}_u$ should be taken over the support of $P(u)$ since
it is not required from policies to find a path in a domains outside support of $P(u)$ and $p^\pi_u$ becomes undefined. 

\footnotetext{Like cost functions in POMDP a number of other functions can be used
instead of \eqref{eq:cost} as long as for each domain $u$ they monotonously increase by $p^\pi_u$.}

Now we can define the navigation problem considered in this paper:
for given $\Omega$ find policy (or its approximation) $\pi^*$ that minimizes cost function \eqref{eq:quality}:
\begin{align}
	\pi^* = \argmin_{\pi}\,\mathbb{E}_u \frac{|p^\pi_u|}{|p^*_u|}
\end{align}

Note that prior probability $P(u)$ implicitly defines structure of space of possible domains or weights. 

\section{Policy model}
The policy definition \eqref{eq:policy} is a very general one since it contains a great class of algorithms. However, optimization in such algorithmic space can be problematic.
Likely, it is possible to narrow this space to mappings from `extended' observations to actions, i.e. to stateless algorithms.

Firstly, only algorithms able to find a path in all possible domains are worth considerations.
Since number of possible observations, robot's positions and directions are finite, only finite number of policy states is used.
An algorithm is ether effectively stateless (can be represented through mapping) or is Pareto dominated by some mapping:
$$\hat{\psi}(\omega^t, s) = (\psi(\omega^t, \hat{s}(\omega^t)), s)$$
where $\omega^t$, $\hat{s}(\omega)$ --- one of possible for $\omega$ states, that minimizes cost, $s$ --- the only state of the modified policy.

Note that definition of $\omega$ as accumulated vision observation is essential for this considerations. For convenience we will define full observation $\hat{\omega}^t$ as tuple $(x^t, y^y, d^t, \omega^t, x^f, y^f)$ and will denote it simply as observation $\omega^t$ in text below.
Correspondingly we extends observation set $\mathcal{O}$.

Following the approach \cite{whitbrook2008genetic} we define navigation policy as set of genes $G$, where each gene from $G$ is a condition-action pair $(c, a)$, where $c: \mathcal{O} \mapsto [-1, 1]$ and $a \in \mathcal{A}$. Conditions $c$ can be viewed as predicates of fuzzy logic with operators described below, where $-1$ corresponds to `false', $1$ to `true' and $0$ to maximal uncertainty.

The final action can be form in different ways, we consider only two of them:
\begin{align}
	(\cdot, \pi^G_1(\omega)) = \argmax_{g = (c, a) \in G} w(g) c(\omega)
\end{align}
\begin{align}
    \pi^G_2(\omega) = \argmax_{a \in \mathcal{A}}
    \frac{\sum_{g = (c, a')} w(g) c(\omega) \mathbb{I}[a' = a \wedge c(\omega) > 0]}
    {\sum_{g = (c, a')} w(g)\mathbb{I}[a' = a \wedge c(\omega) > 0]}
\end{align}
where:
$$I[p] = \begin{cases}
	1,& \mathrm{if}\ p;\\
    0,& \mathrm{otherwise}
\end{cases}$$ and $w(g) > 0$ --- weight of gene $g$, which will be used later for reinforcement learning.
We will refer to $\pi^G_1$ or $\pi^G_2$ as to fusion rules.

Conditions are formed from a predefined finite set of predicates $B$.
Each predicate from $B$ like conditions is a function $\mathcal{O} \mapsto [-1, 1]$.
Let's order predicates of $B$ into vector of basic predicates $b$ and if size of $B$ is $n$ and $\alpha \in \mathcal{R}^n$ then the convolution operator is defined as:
\begin{align}
	(\alpha \otimes b)(\omega) = \sum^{n}_{i = 1} \frac{\alpha_i b_i(\omega)}{\|\alpha\|_1}
\end{align}
where $\|\alpha\|_1$ --- $l_1$ norm.

Set $B$ defines genetic model $\mathcal{G}$:
\begin{align*}
	\mathcal{G} = \left\{ \left( \alpha \otimes b, a\right) \mid \alpha \in \mathbb{R}^n,\, a \in \mathcal{A} \right\}
\end{align*}
Hence within genetic model each gene can be encoded by real vector and action.

The most basic set of predicates $B_0$ consists of elementary vision predicates, position predicates, direction predicates and goal predicates. If $\omega = (v, p, d, g)$ then these predicates can be expressed as:
\begin{align*}
	b^v_{x, y}(\omega) &= v(x, y);\\
    b^p_{x}(\omega) &= \frac{p_x - x}{W};\\
    b^p_{y}(\omega) &= \frac{p_y - y}{H};\\
    b^d_{d'}(\omega) &= d \cdot d';
\end{align*}
where $d \cdot d'$ --- scalar product. Goal predicates $b^g_{x}$ and $b^g_{y}$ are similar to position predicates $b^p_{x}$, $b^p_{y}$.

It easy to see, that for each observation $\omega$ model $\mathcal{G}_0$ contains a condition $c_{\omega}$, that distinguishes $\omega$:
$$\omega=\argmax_{\omega' \in \mathcal{O}} c(\omega')$$
which means, that, using fusion rule $\pi^G_1$, any policy can be represented with some $G \subset \mathcal{G}_0$. However the same is not always possible for $\pi^G_2$\footnotemark.
\footnotetext{However, it is quite easy to build a model that makes every policy expressible in this model with fusion rule $\pi^G_2$.}

The main informal assumption for this model is that in spite of the fact that in general $2^{H \times W} \times (W \times H)^2 \times 4$ genes is required to express some policy, the efficient ones contains large clusters of rules, that can be efficiently expressed by one gene,
reducing size of $G$ to computationally acceptable number, and also that policies build with fusion rules $\pi^G_2$ contains efficient ones.

Unlikely, experiment shows that the assumption may not be true for model $\mathcal{G}_0$ --- none of the obtained policies formed by $\mathcal{G}_0$ were able to find solution in at least one domain.
The problem lies in initial guess for search --- it's very unlikely to
randomly generate acceptable policy to start search with, and unable to evaluate policy search algorithms degenerates into random search.

To solve this problem we extend model $\mathcal{G}_0$ with heuristics, which makes efficient algorithms be encoded by smaller sets of genes, at least for some domain probability spaces $\Omega$.

The model can be extended by custom predicates $\mathcal{O} \mapsto [-1, 1]$, for example, euclidean distance from current position $p$ to the goal $g$:
\begin{align}
	b^r(\omega) = 1 - 2 \frac{\| p - g \|_2}{\sqrt{ H^2 + W^2 }}
\end{align}
where $\|\cdot\|_2$ --- euclidean norm;

It is also noticeable that `predictive' predicates improve quality of the result policies, for example, `will forward command increase real distance to the goal' with possible values $1$ --- it is strictly guarantied, $0$ --- unknown and $-1$ --- the opposite of $1$.

But the major changes in structure of optimal policies are introduced by complete navigation heuristics, obtained from other policies, which might not be optimal, but able to find a path to goal point in every domain.
If a complete stateless policy is denoted as $\pi_0: \mathcal{O} \mapsto \mathcal{A}$, it produces $|\mathcal{A}|$ predicates $b^a$ for each $a \in \mathcal{A}$:
\begin{align}
	b^a(\omega) = 2 \mathbb{I}[ \pi_0(\omega) = a] - 1
\end{align}

In the experiment for such heuristic we chose `naive' navigation algorithm, which follows the shortest path from current point to the goal considering unknown points as free.
At one hand, it is computationally cheap and requires $O(H \times W)$ operations or fewer depending on domain structure and implementation of the algorithm. At the other hand, it is good approximation for optimal policies for domain probability spaces with wide support and high entropy (i.e. under wide assumptions).

Since now model contains complete policy that can be expressed with only 3 genes (that distinguishes 3 possible outcomes of heuristic), other rules are ether useless, since they cover special cases of observation which are already covered by these 3 basic genes, or play role of exceptions from this complete policy which makes more policies evaluable which allows to perform search without degenerating into random guessing.

\section{Policy learning}
Since the model naturally suits into terms of genetic algorithms and due to high dimensionality, we use simplification of Learning Classifier System \cite{farmer1986immune} similar to \cite{whitbrook2008genetic}.

The standard for real vectors genetic procedures such as generation, mutation and crossover are used.

Selection is performed by changing weight $w(g)$ of gene $g$ during learning phase. Genes with negative or low\footnotemark weight are removed from set $G$. Each performed action is evaluated and reward $R \in \mathbb{R}$ spreads as increase in weight for  all genes responsible for performed action. 

For the fusion rule $\pi^G_0$ only the gene $\hat{g}$ responsible for the current action receives penalty or reward:
$$w^t(\hat{g}) = w^{t - 1}(\hat{g}) + R$$

For the other fusion rule reward is shared by all genes, that `voted' for performed action. If gene $g = (c, a)$, observation on step $t$ was $\omega$, and performed action was $a'$ then for each gene from $G$:
$$w^t(g) = w^{t - 1}(g) + \frac{R}{Z(a')} c(\omega) \mathbb{I}[a' = a \wedge c(w) > 0.]$$
where:
$$Z(a') = \sum_{g \in G, g = (c, a)} c(\omega) \mathbb{I}[a' = a \wedge c(w) > 0]$$

\footnotetext{For example, used in the experiment strategy virtually divides set $G$ into active zone and `incubation' zone with restricted size $N$ of the first one. After each step top $N$ genes achieved predefined minimal life time are selected by weight.}

Evaluation of each gene performance is built on decrease of potential between states before and after performing an action: $R = \beta \Delta C$ where coefficient $\beta$ regulates speed of learning.

Depending on the way potential is defined, there are two types of learning: supervised, when potential directly corresponds to the true cost function for current domain and so uses unknown for the robot information about the domain, or unsupervised, when some heuristics are applied to approximated cost function.
For supervised learning if state (position and direction) before performing an action is $s$ and the one after --- $s'$ then:
\begin{align}
	\Delta C = \frac{\rho(s) - \rho(s')}{\rho(s_0)}
\end{align}
where $\rho(s)$ --- length of the shortest path from state $s$ to the goal point, $s_0$ --- initial state.

For unsupervised learning expression for the potential is similar, however now distance $\rho$ should be estimated from current observation $\omega$ rather than from known map of the domain:
\begin{align}
	\Delta C = \frac{\rho_\omega(s) - \rho_\omega(s')}{\rho_\omega(s_0)}
\end{align}
Note that `naive' policy follows decrease of the potential, that approximates distance $\rho$ by considering unknown cells free and hence the more cells are known, the more accurate the estimation becomes. But direct usage of this approximation makes `naive' policy the optimal one. Instead, to make estimation more accurate, we suggest usage of observation from `future' to form current estimation by holding gene evaluation for some number of steps $M$.
\begin{align}
	\Delta C^t = \frac{\rho_{\omega^{t + M}}(s) - \rho_{\omega^{t + M}}(s')}{\rho_{\omega^{t + M}}(s_0)}
\end{align}

It's reasonable to start with low $M$ in order to achieve greater learning speed, gradually increasing $M$ as policy becomes stable.

However, even for big enough $M$ the procedure does not guarantee the same result as supervised learning, however it is still able to correct policy in some situations, for example, penalize for entering dead-end.

\section{Experiment}
For the experiment 20 office-like domains with size about 100 by 100 cells were selected.
These domains form support of probability space $\Omega$ with equal probability of each domain. Starting and goal points are placed on the opposite sides: $p_{0, y} = 0$, $g_{y} = H - 1$. Vision radius was set to $r = 5$.
`Naive' policy $\pi_0$ alone shows $C(\pi_0) \approx 1.8$ and depending on particular domain $u$: $C(p^{\pi_0}_u) \in [1.7, 2.0]$.
Since fusion rule $\pi^G_2$ simultaneously changes weights of group of genes, it was selected for the experiment.
Experiment model extends $\mathcal{G}_0$ by `naive' policy, dead-end detection, direct distance to the goal point, detection of obstacles on right, left sides and ahead of the robot within  vision radius.

\begin{figure*}[H]
  \centering
  \begin{subfigure}[b]{0.3 \textwidth}
  \centering
    \includegraphics[width=\textwidth]{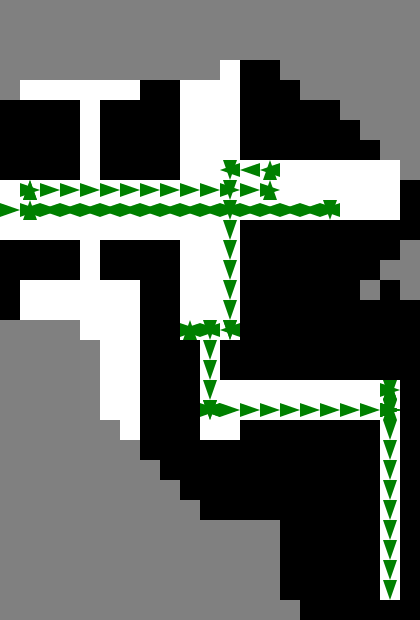}
    \subcaption{First generation.}
  \end{subfigure}
  ~
  \begin{subfigure}[b]{0.3 \textwidth}
  \centering
    \includegraphics[width=\textwidth]{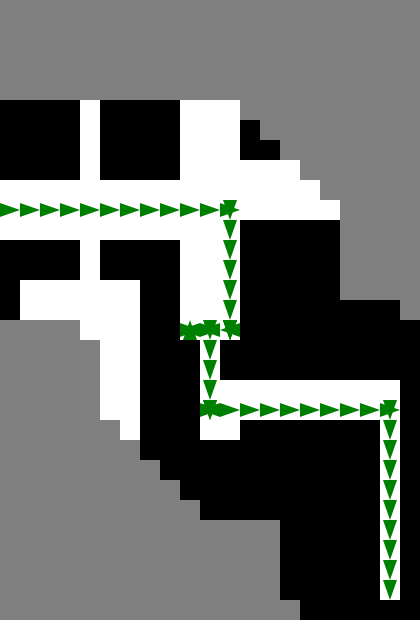}
    \subcaption{10-th generation.}
  \end{subfigure}
    ~
   \begin{subfigure}[b]{0.3 \textwidth}
   \centering
     \includegraphics[width=\textwidth]{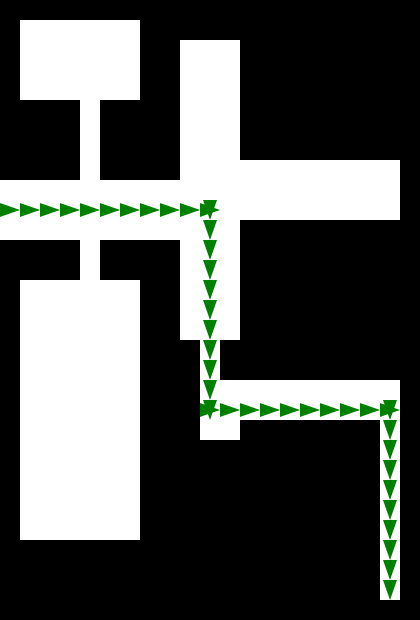}
     \subcaption{The shortest path.}
   \end{subfigure}
   \caption{Demonstration of convergences. Support of the domain probability space consists only from one domain. Triangles represent positions and directions of the robot on its path.}
\end{figure*}

\begin{figure*}[H]
  \centering
  \includegraphics[width=\textwidth]{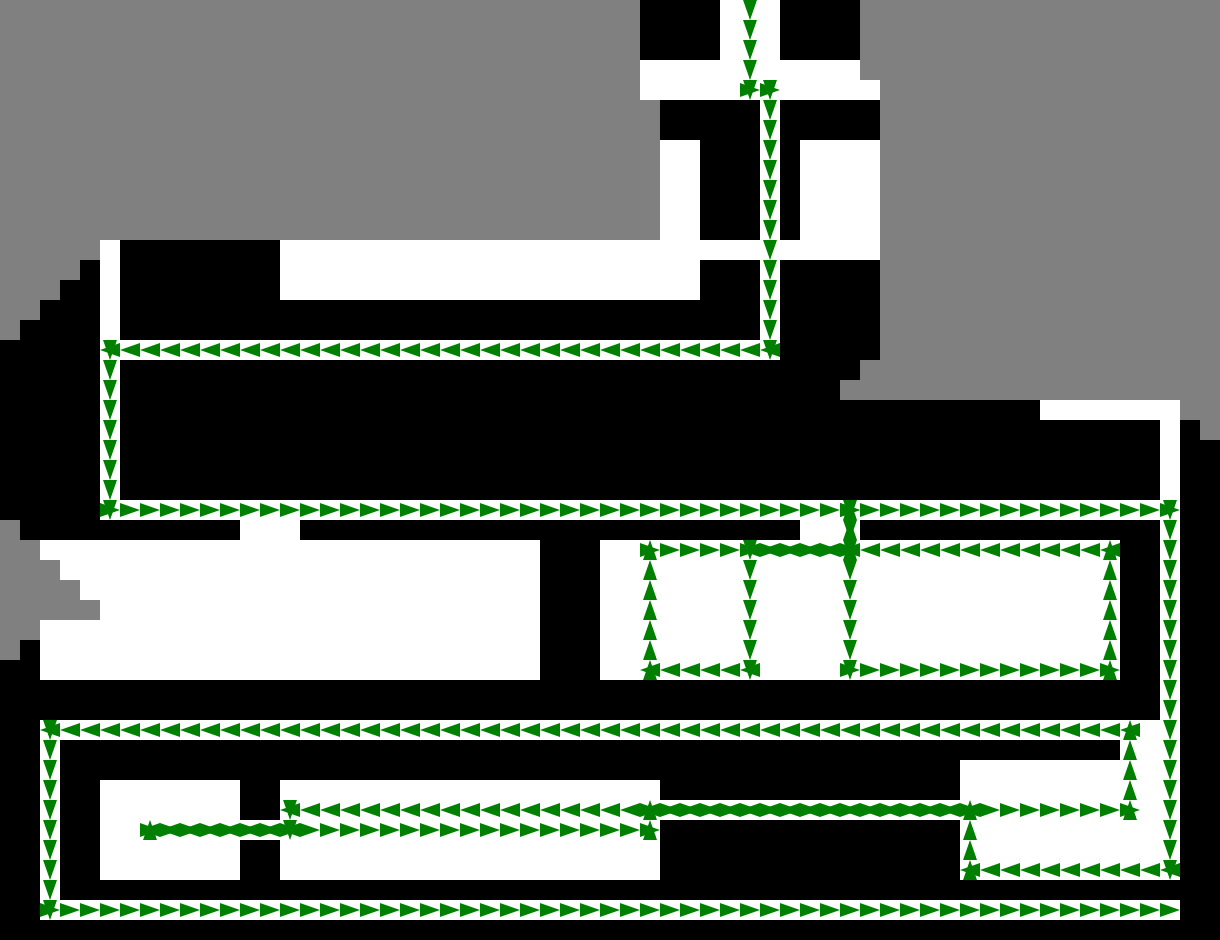}
  \caption{An example of domain from training set in the experiment. Path of `naive' policy is shown.}
    \label{fig:domianexample}
\end{figure*}

\begin{figure*}[H]
  \centering
    \includegraphics[width=\textwidth]{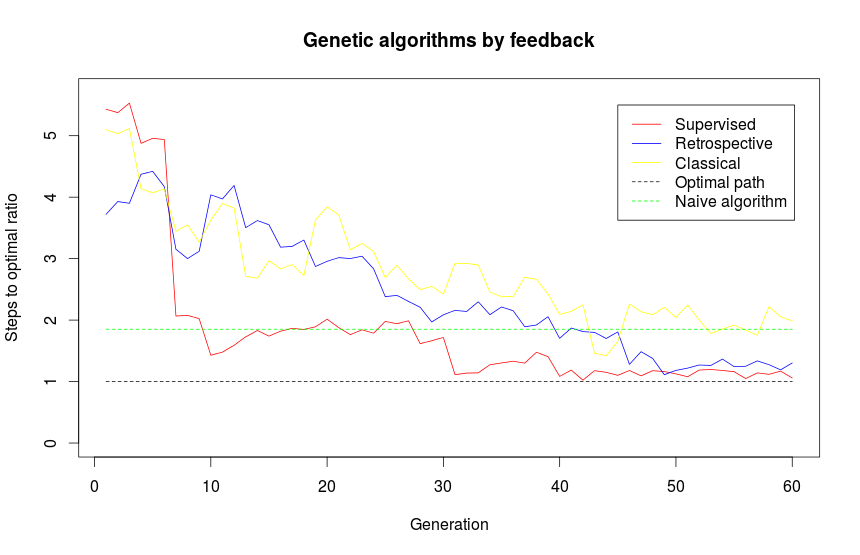}
  \caption{
  	Vertical axis corresponds to the cost function, horizontal axis --- generation number. The training set consists of only one domain shown on figure \ref{fig:domianexample}.
  }
    \label{fig:single}
\end{figure*}

\begin{figure*}[H]
  \centering
      \includegraphics[width=\textwidth]{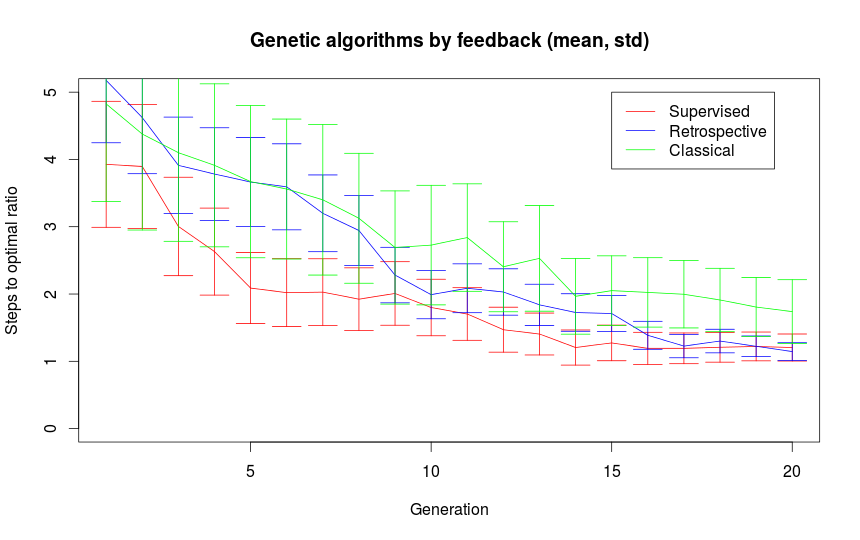}
      \caption{Experiment results. Bars represent standard deviation within 20 selected domains.}
            \label{fig:meand_std}
\end{figure*}

The robot placed into each of 20 domains one by one in random order. The same process have been repeated until convergence is achieved. Iteration of all 20 domains is denoted as generation.

Three different potentials were compared: supervised, unsupervised (denoted as retrospective) with increasing step by 2 each generation starting with $M = 0$ and potential of `naive' policy denoted as classical (unsupervised with $M = 0$).

As the result of the experiment policies with cost $C(\pi) \approx 1.4, 1.5$ were obtained. As it can be seen from the figure \ref{fig:meand_std} in the experiment unsupervised learning has lower convergence speed, but the same cost limit, in contrast to `naive' policy potential, which converges around cost of `naive' policy as predicted.

\section{Future work}
As the experiment shows obtained policies widely use heuristic predicates. Possibly, forming proper predicate dictionary, it is achievable to fully negotiate model $\mathcal{G}_0$, or at least, introduce two types of genes: standard and formed exclusively from heuristic predicates, with majority of the last type. Such dictionary allows to considerably reduce number of optimization parameters, which may speed up optimization or even allow usage of classical method of discrete optimization.

\bibliographystyle{aaai}
\bibliography{bibliography}

\end{document}